# A Review on Near-Duplicate Detection of Images using Computer Vision Techniques

K. K. Thyagharajan,
kkthyagharajan@yahoo.com
Professor & Dean (Academic)
RMD Engineering College
Tamil Nadu, INDIA

G. Kalaiarasi
kalaikannan.l@gmail.com
Assistant Professor
Sathyabama Institute of Science and Technology
Chennai, INDIA

## Abstract

Nowadays, digital content is widespread and simply redistributable, either lawfully or unlawfully. For example, after images are posted on the internet, other web users can modify them and then repost their versions, thereby generating near-duplicate images. The presence of near-duplicates affects the performance of the search engines critically. Computer vision is concerned with the automatic extraction, analysis and understanding of useful information from digital images. The main application of computer vision is image understanding. There are several tasks in image understanding such as feature extraction, object detection, object recognition, image cleaning, image transformation, etc. There is no proper survey in literature related to near duplicate detection of images. In this paper, we review the state-of-the-art computer vision-based approaches and feature extraction methods for the detection of near duplicate images. We also discuss the main challenges in this field and how other researchers addressed those challenges. This review provides research directions to the fellow researchers who are interested to work in this field.

# 1 Introduction

There is a huge number of digital images on the web and it is common to see multiple versions of the same image. The increasing use of low-cost imaging devices and the availability of large databases of digital photos and movies makes the retrieval of digital media a frequent activity for a number of applications. Creation, display and management of digital photos have been important activities in the digital life. People are accustomed to recording their daily life or journeys by digital cameras and share their living/travel experience on the web. Handling images on the Internet is very easy and convenient for users nowadays. This brings numerous security problems. Social media is one of the main areas where massive propagation of similar visual content takes place. The manipulation of images using computer techniques is not new and gained a lot of popularity and even acceptance in diverse areas such as forensic investigation, information technology, intelligence services, medical imaging, journalism, digital cinema, special effects in movies, etc. As more images are published on the web, modifying images is becoming increasingly easy with the help of powerful and user-friendly image manipulation software, and the move towards paperless workplaces and the introduction of e-Government services demand more data to be stored in digital format and more challenges are to be faced to securing authentic data. Unfortunately, document files, voice data and image data are all vulnerable to manipulation and doctoring. This makes the slightly altered image copies, which are termed near-duplicate (ND) images to their originals, difficult to detect. Proper feature extraction methods should be employed to overcome this challenge. Feature extraction is the most important step in near-duplicate detection. It is most critical because the features extracted directly influence the efficiency in the detection/retrieval.

The presence of near-duplicates (NDs) plays an important role in the performance degradation while integrating information from different sources. Web faces huge problems due to the existence of such NDs. By introducing efficient methods to detect and remove such NDs from the web not only decreases the computation time but also increases the relevancy of search results. The near duplicates must be clustered to avoid viewing the re-occurrence of the same image or variants of it resulting in efficient search engine results. Near-duplicate images arise generally when the same object is captured with the different viewpoint, illumination conditions, and different resolutions with zooming effects. These images are also called similar images because by definition [1], similar images are the images of the same object or scene viewed under different imaging conditions. Figure 1 shows an example of the near-duplicate image taken from the INRIA Copy days dataset [1]. The Fig. 1b is JPEG attacked image of Fig. 1a. Both look visually similar, but there exists difference.

Near-duplicate images also arise in the real life when a particular object is captured at different instances of time when the background is moving as in panorama and burst shots. The examples taken from California ND data set [2] are shown in Fig. 2.

Panorama photos capture the images with the horizontally elongated view and it is also known as wide format photography. Burst shots are useful for capturing the perfect moment when the objects in the image are moving.

Figure 3 shows the example of near-duplicate/forged/tampered images taken from the COVERAGE database [3]. In Fig. 3, a) is the original image and b) is the tampered version of a). Both look visually similar, but there exists difference. The portion marked in green color is chosen as the region of interest and the lighting of this region is modified (illumination transformation), translated to another similar object of the same image.

Two NDIs may be perceptually identical, in which the differences comprise of noise editing operations, small photometric distortions, change in color balance, change in brightness, compression artifacts, contrast adjustment, rotation, cropping, filtering, scaling etc. Figure 4 shows two images taken from Toys dataset [4] as an example of near duplicates for rotation transformation. Figure 4b) is the rotated image of Fig. 4a). Figure 5 shows the images taken from California ND dataset [2] as an example of near duplicate image captured from the same scene.

The presence of these ND images can be detected using Computer Vision techniques. Computer Vision deals with how computers can be made to gain high-level understanding from digital images or videos. It seeks to automate tasks that the human visual system can do. Computer vision tasks include methods for acquiring digital images, image processing and image analysis to reach an understanding of digital images. The problem in Computer Vision, image processing and machine vision is that of determining whether or not the image data contains specific object, feature or activity. In Computer Vision, a feature is a piece of information which is relevant for solving the computational task related to certain application. The feature concept is very general and the choice of features in a particular computer vision system may be highly dependent on the specific problem. Features are used as a starting point for many computer vision algorithms. Since features are used as the starting point and as it is considered as main primitive, the overall algorithm will only be as effective as its feature extraction. So, this survey concentrates mainly on the feature extraction and how it is efficiently used in the detection of NDIs.

## 1.1. Motivation

Detecting near duplicate images (NDIs) in large databases should meet two challenging constraints—i) For a given image, only a small amount of data can be stored in web to reduce the search complexity (for e.g. fingerprint) and ii) Queries must be very simple to evaluate [5]. Image retrieval from large databases is a fundamental issue for forensic tasks. Forensic tasks include collecting, identifying, classifying and analyzing physical evidence related to criminal investigations. Since the images on the web are retrieved easily, it becomes easy for the web users to manipulate many images as near duplicates. Mainly the images that are used as physical evidence in the criminal investigation are manipulated. The presence of NDs plays an important role in the performance degradation while integrating information from different sources. Web faces huge problems due to the existence of such NDs. This increases the index storage space and thereby increases the serving cost. By introducing efficient methods to detect and remove such NDs from the web search not only decreases the computation time but also increases the relevancy of search results. Some of the applications of near-duplicate detection (NDD) include integrating TV news telecasted on different days, integrating information from various letter sorting machines for postal automation, etc.

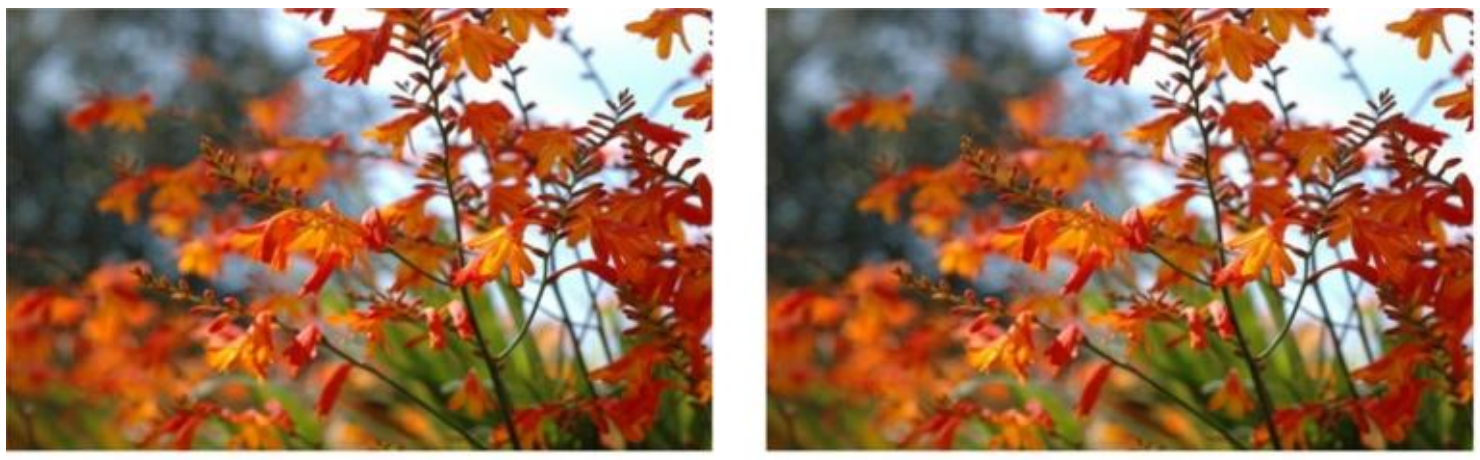

(a) Original Image (b) JPEG Attacked Near-Duplicate Image

**Fig. 1** Examples of near duplicate images from INRIA Copydays dataset

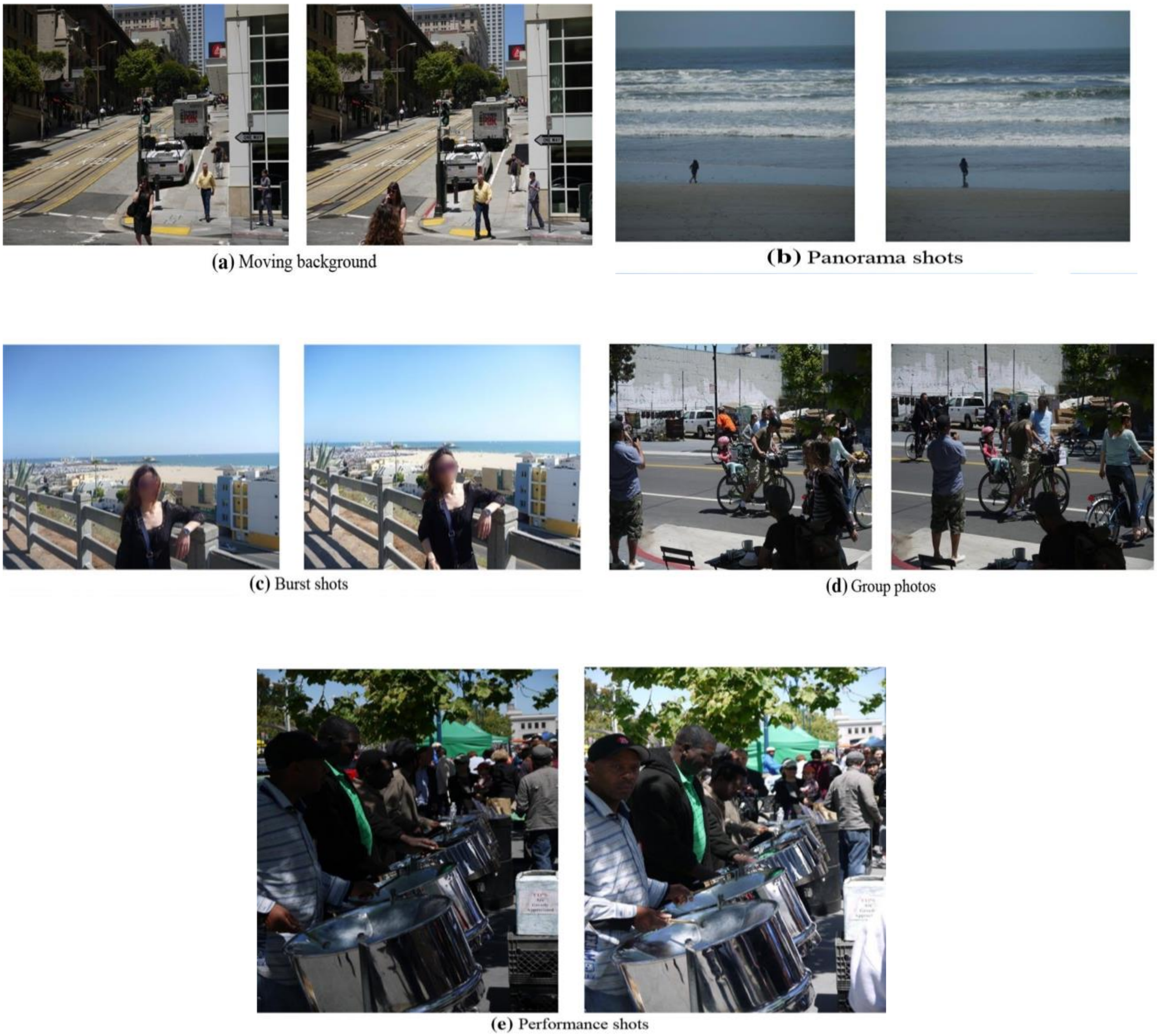

**Fig. 2** Examples of near-duplicate images from California ND dataset

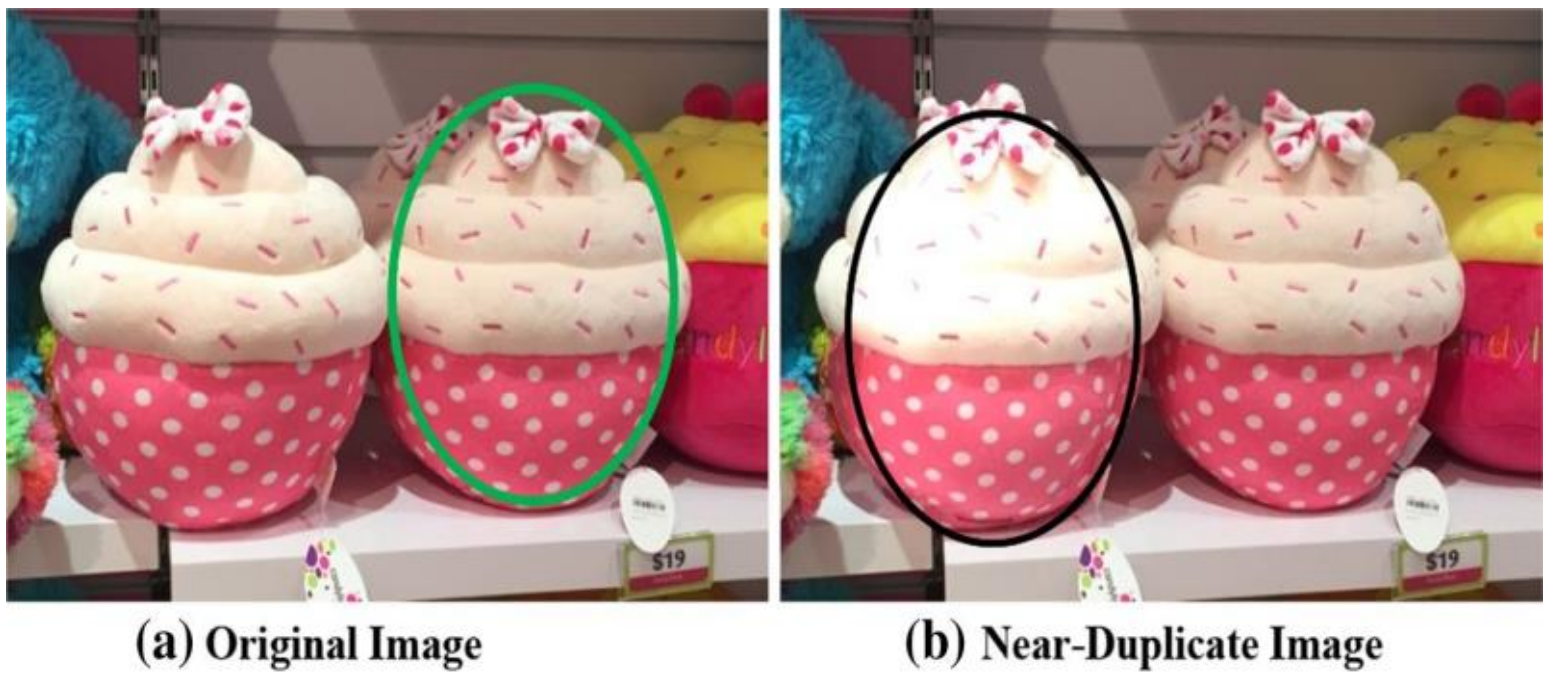

**Fig. 3** Example of near-duplicate image (visually similar)

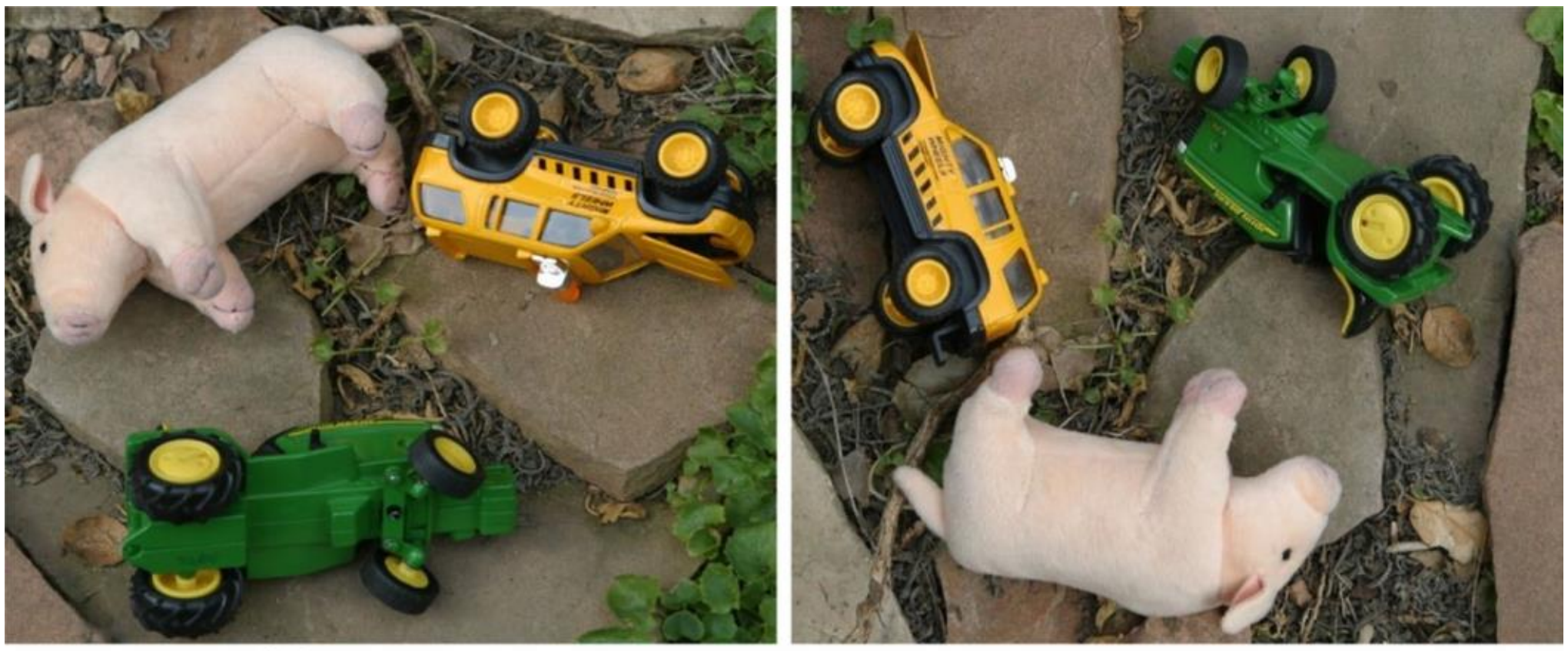

(a) Image (b) rotated image

**Fig. 4** Near-duplicate image created by rotation

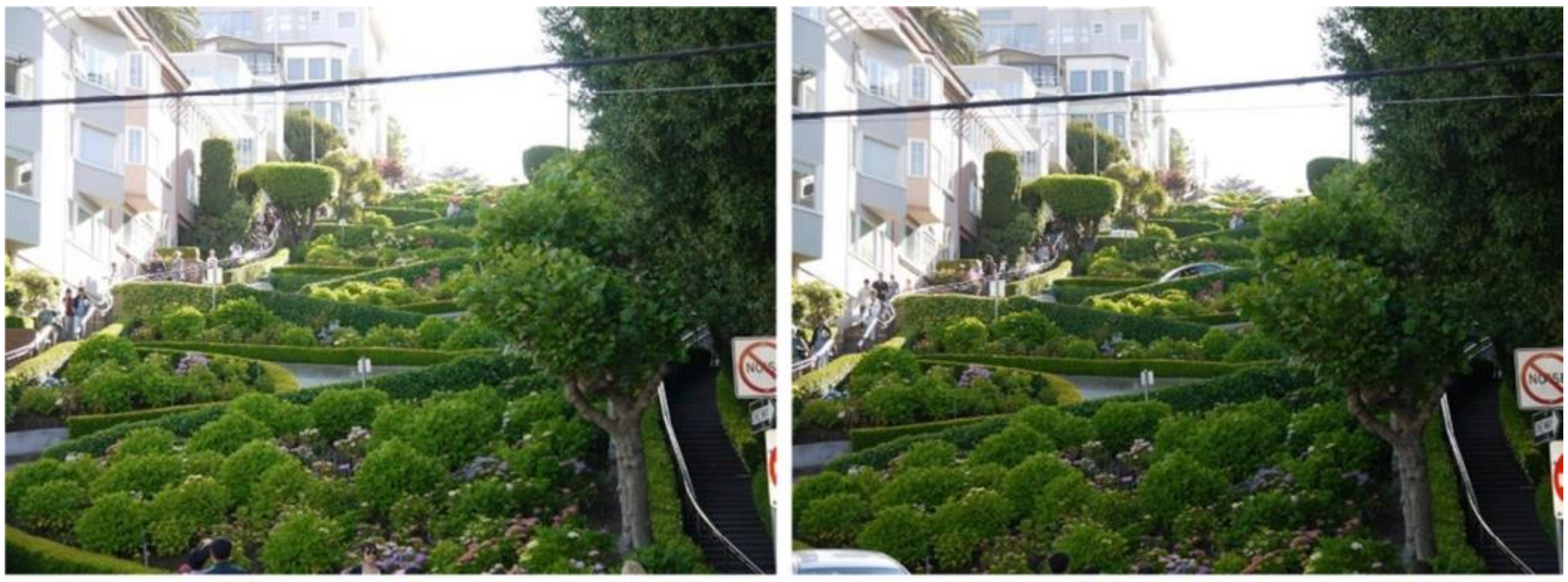

(a) Image 1 (b) Image 2

**Fig. 5** Near-duplicate images with viewpoint and illumination change

Landge Amruta and Pranoti Mane reviewed the development of several image matching algorithms [6]. Several mathematical formulae are discussed based on which the similarity between two images is computed. But, the review does not concentrate on the image matching algorithm that best suits the near-duplicate identification. The state-of-art review paper in copy-move forgery detection concentrates only on one part of the near-duplicate image detection i.e. detecting forged images using image editing software [7]. As forged images are the subset of ND images and they concentrate on detecting the type of forgery made, the datasets requirements to evaluate these methods will be slightly different. Evaggelos Spyrou and Phivos Mylonas made a survey on the Flickr social network highlighting the current progress, innovations, and the evaluation methods [8]. If any two images look similar or convey similar concepts to you they are near-duplicates. So, near-duplicate detection is a highly subjective process. An algorithm which is producing good results for the ND images created using editing software need not produce the same result for the ND images captured using a camera in real life. For example, an artificial zooming effect created using software will not be exactly same as that of the effect produced by zooming lens of a camera and it will not be accurate because while zooming with the camera, there may be slight changes in the scene. So to fill with this gap any algorithm developed for NDD should also be evaluated using datasets created from the real-life scenario. NDD is a broader area of research.

The remainder of this paper is organized as follows. Section 2 reviews the general methods used for the detection of near duplicates. The feature extraction methods for NDD images are reviewed in Sect. 3. In Sect. 4, the methods used in ND retrieval of images are presented. NDD used in image forensics are mentioned in Sect. 5. Section 6 discusses the clustering methods used for the NDD. Some of the databases used for the evaluation of NDD are explained in Sect. 7 and the applications of NDD images are listed in Sect. 8. The Sect. 9 points to the challenges and future research directions and finally, Sect. 10 concludes this paper.

## 2 General Method Used for the Detection of Near Duplicates

The steps followed in the detection of near-duplicate images are generally quite similar to that of traditional image matching method. Figure 6 shows the general block diagram for the identification of near duplicate images. Initially, the features of the images (image descriptors), retrieved from the web or in the database, are extracted and stored in another database named ‘Feature Vector Database’. The features of the query image are extracted and compared with the feature vectors of all images already stored in the database.

If matches are found, they are marked as near-duplicate/ visually similar images. This approach requires an efficient indexing structure to search for similar descriptors in the database. The percentage of similarity required is fixed by a reference value. Some of the methods used for the detection of near-duplicate images are presented in the next section.

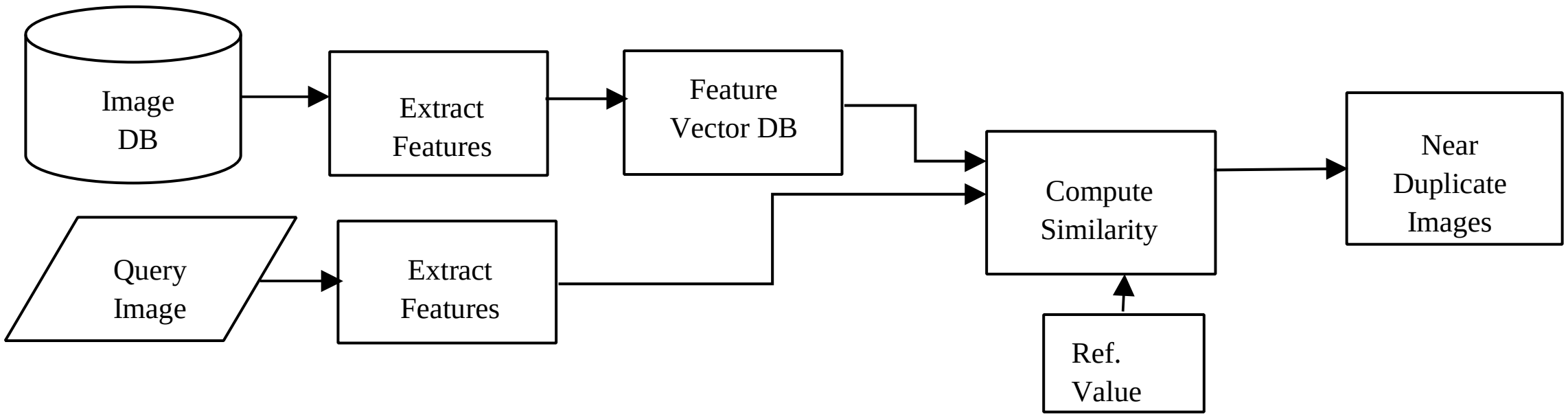

**Fig. 6** General block diagram for identifying near-duplicate images

## 3 Feature Extraction Methods for NDD Images

Generally, features extraction methods can be broadly classified as low-level and high-level features or local and global features. In NDD it can be classified as point-based feature extraction, pixel-based feature extraction and area-based feature extraction methods. If the features are extracted based on the key points or interest point, then those methods are classified as point-based feature extraction methods. If the features are calculated at each and every pixel, then it is referred as pixel-based feature extraction method. Area-based feature extraction methods are the methods in which the features are calculated over the entire image or just regular sub-area of an image. This classification is shown in Fig. 7.

### 3.1 Key point-Based Feature Extraction Methods

The state of art near-duplicate image/video detection methods requires proper selection of image descriptors (features) to represent the images precisely. For finding copyright violations and detecting forged images, Ke et al. presented an efficient near- duplicate detection method by using robust interest point detector (Difference of Gaussian, DoG) and distinctive local descriptors (Principal Component Analysis – Scale Invariant Feature Transform, PCA-SIFT) [9]. They used locality sensitive hashing (LSH) to index these local descriptors. But, the system matches the similar images of the same scene even if they are neither near duplicates nor forged images. Chen Li and Fred Stentiford presented an attention-based similarity measure which is generated based on a trial and error basis [10]. The SIFT interest points are very effective to geometric variations in the image but scalability is a big issue due to a large number of features generated. To overcome this issue, Foo and Sinha pruned SIFT interest points to reduce the memory and query runtime with negligible loss in effectiveness [11]. To reduce the number of SIFT interest keypoints, the threshold is varied to discard candidate local peaks. They also pruned PCA-SIFT features. A modified Redundant Bit Vector (RBV) index is introduced by Foo and Sinha for a high-dimensionality search of multimedia data and is efficient for audio fingerprint detection [12]. RBV gained efficiency and a reduction in memory requirement in comparison to LSH.

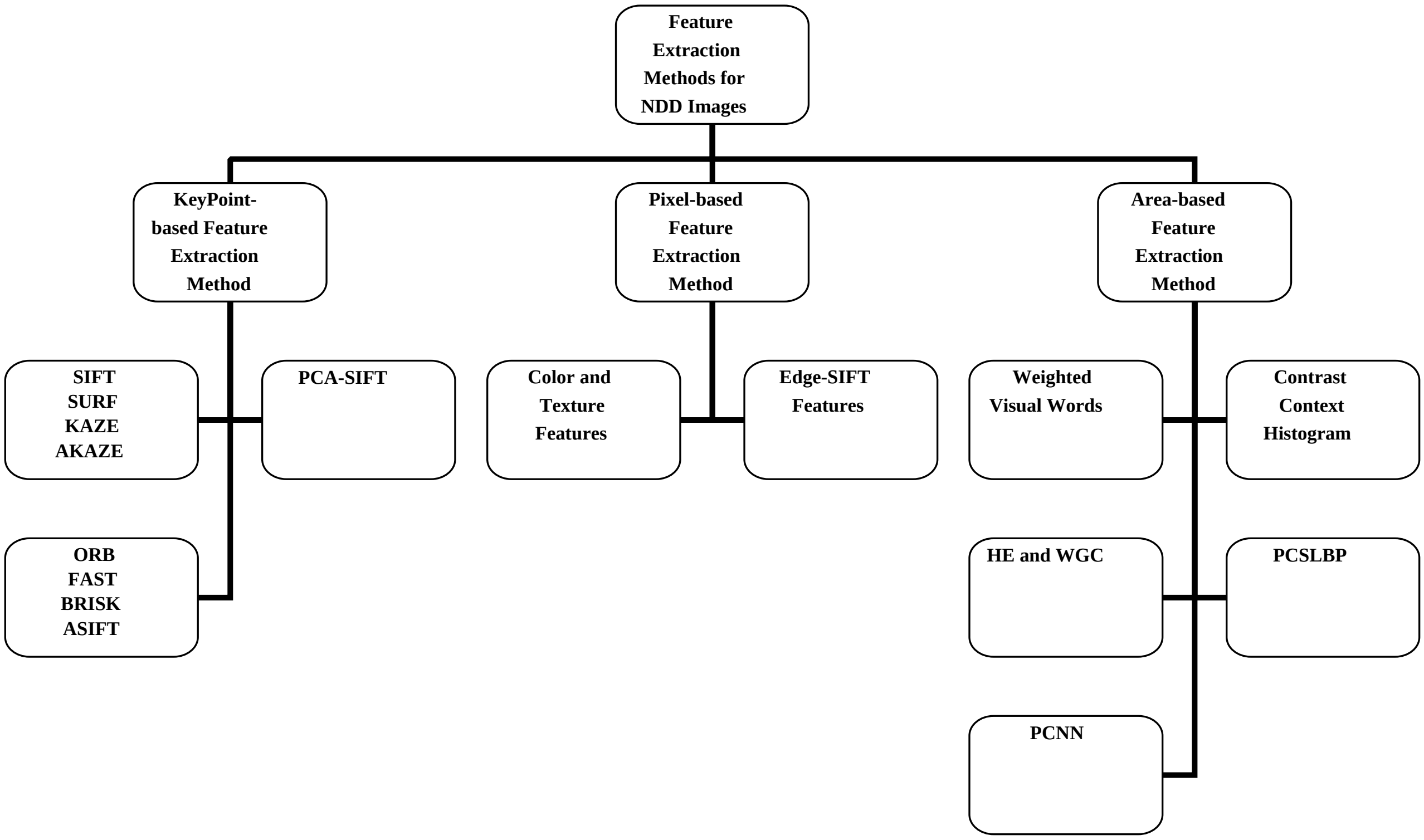

Fig.7. Classification of Feature Extraction Methods used with NDD Images

SURF (Speeded Up Robust Features) is a scale and rotation-invariant detector [13]. It is used for classification tasks and is faster to compute. It is suited for object detection, object recognition or image retrieval. Hessian matrix approximation is used for interest point detection. SURF outperforms SIFT. This is due to the fact that SURF integrates the gradient information within a sub patch, whereas SIFT depends on the orientation of the individual gradients. This makes SURF less sensitive to noise. Another feature ORB (Oriented FAST (Features from Accelerated Segment Test) and Rotated BRIEF (Binary Robust Independent Elementary Features)) is a rotation invariant and resistant noise descriptor [14]. ORB is a combination of oFAST and rBRIEF. BRIEF is a feature descriptor that uses simple binary tests between pixels in a smoothed image patch. Its performance is similar to SIFT in many respects, including robustness to lighting, blur and perspective distortion. However, it is very sensitive to in-planar rotation. FAST features are widely used because of their computational properties. However, FAST features do not have an orientation component. The main drawback in ORB is that scale invariance is not addressed.

SIFT not only has good scale and brightness invariance but also has a certain robustness to affine distortion, perspective change, and additive noise. However, to extract SIFT features for representing an image, hundreds or even thousands of SIFT key points need to be selected. Each key point uses a 128-dimensional feature vector to explain the features. Thus, the matching value of detection methodology supported SIFT options is high. Cao et al. experimented with ASIFT to detect near duplicates [15]. It was found that it showed better performance than Hessian-Affine and MSER (Maximally Stable Extremal Region). ASIFT (Affine Scale Invariant Feature Transform) manages effectively all the parameters of the affine transform [16]. Lindeberg presented a theoretical basis for computing scale invariant image features and image descriptors for a wide range of possible applications in computer vision [17]. Wang et al. presented a keypoint based approach for near duplicate image detection [18]. This method works with fewer keypoints matching, i.e., it can detect reliable and salient keypoints and these keypoints are matched with more strong constraints. The false alarm is reduced as more specific details are not considered to enhance the color matching.

Tareen and Saleem analysed SIFT, SURF, KAZE, AKAZE, ORB and BRISK features [19]. They concluded the following SIFT is invariant to image rotation, scale and limited affine variations; its main drawback is high computational cost. SURF features are invariant to rotation and scale but they have little affine invariance. The main advantage of SURF over SIFT is its computational cost. KAZE features are invariant to rotation, scale, limited affine and have more distinctiveness at varying scales with the cost of moderate increase in computational time. AKAZE Accelerated KAZE. AKAZE features are invariant to scale, rotation, limited affine and have more distinctiveness at varying scales because of non-linear scale spaces.

The ORB features are invariant to scale, rotation and limited affine changes. BRISK (Binary Robust Invariant Scalable Keypoints) features are invariant to scale, rotation and limited affine changes. The overall accuracy of SIFT and BRISK is found to be highest for all types of geometric transformations and SIFT is concluded as the most accurate algorithm. BRISK is at second position with respect to the accuracy for scale and rotation changes. ORB is less accurate than BRISK for scale and rotation changes but both are comparable for affine changes. AKAZE is more accurate than KAZE for scale and affine variations. However for rotation changes KAZE is more accurate than AKAZE. The advantages, disadvantages and suitability of point-based features are tabulated in Table 1.

### 3.2 Pixel-Based Feature Extraction Methods

Foo et al. presented detection of ND images in the web search [20]. The color and texture features are extracted using Dynamic Partial Functions (DPF) and PCA-SIFT local descriptors. They used hash based probabilistic counting (HPC) for automatic detection of ND images. With the fast development of RISC (Reduced Instruction Set Computer) processor, camera, display technology and wireless networks, the mobile devices have become more powerful, ubiquitous and important to users. A large-scale partial-duplicate image search system for mobile platform was developed in [21].

**Table 1** Summaryoffeaturesinpoint-basedfeatureextractionmethods

| Featuresused | Advantages | Disadvantages | SuitabilityforNDD |
|---|---|---|---|
| SIFT[ 10, 11] | scaleandbrightnessinvariancefeaturewithcertain robustnesstoaffinedistortion,perspectivechangeand additivenoise | Matchingcostinthedetectionmethodishighsince thousandsofkeypointsareselected | Suitableforimagematchingwhereminimalmatchingis utilized |
| PCA-SIFT[ 9] | NumberofSIFTinterestkeypointsarereducedthereby reducingthememoryrequirementandqueryevaluation time | SimilarimageofthesamesceneismatchedasNDeven iftheyareneitherNDsorforgedimagesleadingto slightlossineffectiveness | SuitableforAudiofingerprintdetection |
| ASIFT [15] | ASIFTisinvarianttoaffineparameterssuchastransla - tion,zoom,rotationandtwocameraaxisorientation suchaslatitudeandlongitudeangles | ASIFTisslowbecauseofthehighestfeatureextraction time | Canbeusedwheredetectionspeedisnotimportantbut accuracyismoreimportant |

Compared with traditional methods, image search on mobile platform needs to take more factors into consideration, e.g., the limited computational resource, storage and memory capacity, relatively expensive wireless data transmission, rich sensors like GPS, accelerometer, gyroscope etc. To achieve efficient and accurate partial-duplicate mobile search, they focussed on extracting compact, discriminative, and efficient local descriptor which is commonly known as a basis for Bag-of-visual words (BOWs) representation. Edge-SIFT is partially based on SIFT but is more discriminative, compact and suitable for mobile partial-duplicate image search. Edge-SIFT recorded the locations and orientations of edge pixels, thus preserved rich spatial clues and imposed more strict restrictions on feature matching.

Pulse Coupled Neural Network (PCNN) is used in the ND image detection. It is used for feature extraction process as it is simple and extracts reduced number of features. Many researchers work on PCNN for its easy implementation, higher recognition, anti-noise disturbance, robustness, etc. PCNN is unsupervised and self-organized network. The PCNN and its numerous variations have been found to be useful in a wide variety of applications. The mathematical model using discrete Fourier transform on the global pulse signal of the PCNN is described by Raul C. Muresan [22] and analysed the pulse of the network to achieve scale and translation independent recognition of an image. The system is used for recognising simple geometric shapes and letters. Yu and Le presented a novel feature extraction method for image processing via PCNN with Tsallis entropy [23]. This feature is translation and scale independent, while rotation is a bit weak at diagonal angles of 45º and 135º caused by pixel discretization. It is a new and potential approach and can be applied to all sorts of recognition fields. The drawbacks encountered are low classification rate and need to simplify the procedures in this model.

Yide Ma et al. extracted image features as time signals of PCNN [24]. These signals are invariant to larger changes in rotation, scale and shift or skew of the input image. They compute Mean Square Error (MSE) between the feature vectors to decide the kind of image group that the input image belongs. This method is strongly flexible to resist noises and greatly robust to recognise the image. Feature extraction using Unit-linking PCNN was proposed to generate the global and the local image icons as image features [25]. He included not only the intensities or color information but also the geometry structures or color distribution of the images. The global unit-linking PCNN icon is translation, rotation and scale invariant. Local unit-linking PCNN image icon can authenticate images correctly. For a complicated object detection system, the global unit-linking image icon method should work together with other methods to improve the detection performance. Radoslav Forgac and Igor Mokris improved the feature generation using an optimised PCNN [26]. Their main aim was to reduce the number of generated features and to determine the optimum number of iterations. This reduced feature is considered as a feature with maximal information value for image recognition process. Trong-Thuc et al. presented Ram-based and pipeline-based hardware architectures for PCNN for real-time image feature extraction [27]. The advantages of these models are antinoise and invariant against geometrical changes. Mona Mahrous Mohammed et al. proposed an optimized PCNN that extracts the visual features of the images as signatures and uses those features for classification [28]. The key feature of using image signature is its simple and small representation. This signature is represented as a vector that contains the number of firing neurons over several iterations. An et al. proposed a graph matching with geometric constraints for near duplicate image retrieval, which explored the spatial relationship in image patches [29]. Also, introduced geometric constraints to remove the false matches and yielded good retrieval results. It has complex run-time and the index structure is not efficient. Along with PCNN, CNN (Convolutional Neural Network) is also used in the near duplicate detection. Zhang et al. presented an image pair-wise similarity joining learning function without any handcrafted features [30]. Convolutional neural network is used to encode the features and classification. This method processes the raw images directly without human designed feature extraction and the images are classified. Compared to conventional approaches, this method eliminated the complex handcrafted features extraction and processed images jointly.

Vonikakis et al. presented non-linear version of the DoG detector (nLDoG-1) [31]. The nLDoG-1 exhibited increased sensitivity to low contrast and provided detection robustness in low local contrast occasion. The nLDoG-1 performed better than the other existing detectors in terms of repeatability score and number of corresponding regions in sequences with either uniform or non-uniform illumination. It is computationally inexpensive and required fewer memory resources. The detector is suitable for mobile robotics applications and for outdoor or space scenarios, where severe illumination changes occur. This detector is applicable only for uniform illumination changes and not for non-uniform illumination changes. Also, it cannot be used in realistic lighting condition. The same authors also presented an illumination invariant operator by combining the non-linear characteristics of biological center-surround cells with the classic difference of Gaussians operator [32]. The advantage is keypoints are detected in both dark and bright image regions. It is appropriate for outdoor vision systems working in environments under uncontrolled illumination conditions. Zhuang et al. proposed two methods Intensity Difference Quantization and Weakly Spatial Context Coding to improve the discriminative power and robustness of binary descriptors [33]. The advantages, disadvantages and suitability of pixelbased features are tabulated in Table 2.

**Table 2** Summary of Features in Pixel-based Feature Extraction Method

| Features used | Advantages | Disadvantages | Suitability for NDD |
|---|---|---|---|
| Color and texture features [20] | High accuracy | These features will not be able to detect complexity in NDD | Suitable only for detecting the similar pattern of images |
| Edge SIFT [21] | It is more discriminative and compact | High computational complexity | Suitable for mobile near-duplicate image search |
| PCNN [22–28] | Can produce a minimal number of features even for large images. | Difficulty in estimating the optimum values for its large number of parameters | To extract features of NDD images |

### 3.3 Area-Based Feature Extraction Methods

Wu et al. presented query-oriented subspace shifting algorithm [34]. The algorithm measures the similarity in various subspaces that are dynamically generated based on the correlation between samples and the query image. The near duplicates are never missed while detecting as these subspaces are query oriented. But their method is not efficient for large-scale NDD. Chum et al. proposed a near-duplicate image detection method using the min-Hash algorithm [35]. The min Hash method stored only a small constant amount of data per page and a complexity for duplicate enumeration that is close to linear in the number of duplicates returned.

A set of weighted visual words are extracted and each visual word is assigned a weight. The weighted histogram intersection is the best similarity measure in both retrieval quality and searches efficiency. The main drawback is some relevant information is not presented in the set of visual words representation. Chun et al. utilized contrast context histogram for image matching and recognition [36]. Their experiments demonstrated that contrast-based local descriptors represented local regions with more compact histogram bins. Their method is used in many real-time applications as it has high matching accuracy and efficient computation.

Herve Jegou et al. proposed to use Hamming embedding (HE) and weak geometric consistency constraints (WGC) descriptors for image matching and searching in large image datasets [1]. The HE provides binary signatures to refine image matching and WGC filters the descriptors that are invariant to rotation and scaling transformations. Pattern Entropy (PE) is a measure to evaluate the information of being a near duplicate pair. The PE captures the matching patterns with two histograms of matching orientations. PE is excellent for characterizing regions with parallel-like matching lines, but not for regions undergone considerable scale and rotation changes. For a more robust pattern evaluation, scale rotation pattern entropy (SR-PE) is introduced by Zhao and NGO [37]. SR-PE gets a lower entropy value indicating the perfect match only when the ND region pair shows homogeneity across channels. SR-PE is computationally efficient. Compared to PE, SR-PE is capable of evaluating complex patterns composed of ND regions under the unknown scale and rotation changes. SR-PE is not effective when the viewpoint changes and it cannot be used to infer higher-level semantics by localizing object and background duplicates. Battiato et al. exploited the coherence between feature spaces in the image representation and in the codebooks generation to improve the bags of visual phrases [38]. This is achieved through the alignment of the feature space partitions obtained from independent clustering.

Sluzek et al. described the detection and segmentation of ND fragments in random images [39]. Random images are the images with unpredictable content. Such fragments usually represent the identical object, though captured from a different viewpoint, under different photometric conditions and/or by a different camera. Affine transformations poorly model ND fragments for objects with strongly non-planar surfaces or objects which are naturally deformable. For the retrieval of such near duplicates, an alternative algorithm is proposed by Sluzek based on the point-based topological constraints. Topological constraints provide high performances in matching non-linear ND fragments at rather low computational costs. Cho et al. proposed a concentric circle-based image signature for NDD [40]. The image is partitioned by radius and angle levels from the center of the image. The feature values are calculated using the average or variation between the partitioned sub-regions and are formed into an image signature by hash generation. The hashing facilitates storage space reduction and quick matching. The hashed bits are more stable and robust to image modifications. But encountered difficulty in geometrical modifications such as noise addition and blurring which change the center.

The traditional near-duplicate image search systems [41–45] mostly were built on the bag-of-local features (BOF) representation as it is favourable for simplicity and scalability. But it has certain shortcomings – high time complexity of the local feature detection, discriminability reduction of local descriptors due to BOF quantization and neglects the geometric relationships among local features after BOF representation. To overcome these shortcomings, Xie et al. proposed a framework by using graphics processing units (GPU) [46]. The computational capability of GPU is much higher than that of a central processing unit. Due to its powerful capability, the GPU nowadays serves not only for graphics display, but also for general purpose computation such as molecular dynamics, image processing, and machine learning and computer vision. Image matching is one of the key technologies for many visual-based applications including template matching, block motion estimation, video compression, stereo vision, image/video NDD, the similarity join for image/video database and so on. Normalized cross correlation (NCC) is one of the widely used methods for image matching with preferable characteristics such as robustness to intensity offsets and contrast changes, but it is computationally expensive. Satoh proposed features, derived by the method of Lagrange multipliers [47]. By using these features, NCC-based image matching is effectively accelerated. It applies image NDD to 2 billion images in the web for image auto-annotation.

Das et al. proposed the term document weight matrix approach with three phases such as rendering, filtering and verification for finding near duplicates of an input web page from a huge repository [48]. This approach explores the semantic structure, content and context of a web page rather than the content only approach. Accuracy is improved further by applying this method after identifying the main content blocks rather than the entire web page. Dong et al. achieved efficiency and quality in [49] by making three contributions – i) the non-discriminative features are filtered out using entropy-based filtering method and with the retained high-quality features, a single match is sufficient for claiming an ND relationship between images with high confidence. This reduced the number of SIFT features to be indexed, ii) A query expansion method based on graph cut is used to reduce the false positive rate by improving recall, iii) This system is capable of indexing more than 50 million web images on a commodity server and it can return search results in less than two seconds.

Bueno et al. proposed the Bayesian approach for near duplicate image detection and investigated the performance in different probabilistic models [50]. The task of identifying an image whose metadata are missing is applied in different applications such as metadata retrieval in cultural institutions, detection of copyright violations, investigation of latent cross-links in archives and libraries, duplicate elimination in storage management, etc. The increasing interest in archiving all of humankind's cultural artifacts has resulted within the digitization of lots of books. Most of the data found in historical manuscripts are mainly text, but with a few numbers of images. Rakthanmanon et al. introduced a scalable system that can detect approximately repeated occurrences of shape patterns both within and between historical texts [51]. This ability to find repeated shapes allowed automatic annotation of manuscripts and allowed users to trace the evolution of ideas. Li et al. proposed a scheme by combining the neighbourhood information of single local feature and the global geometric consistency of multi-local features for improving the accuracy of BOW model [52]. First, the geometric contextual information of image local features is constructed to enhance the distinctiveness of visual word. Then, the global geometric consistency of subset of-features is verified for improving the accuracy of retrieval results. Li and Feng proposed an approach to detect near duplicate images automatically based on visual word model [53]. SIFT descriptors are utilized to represent image visual content to detect local features of images. NDI detecting process is implemented by histogram distance computing.

In the traditional methods, the Bag-of-Visual Words (BOVW) model and the inverted index structure are used for image matching. Despite the simplicity, efficiency and scalability, these algorithms highly depend on the accurate matching of local features. This suffered from unsatisfied precision and recall. Xie et al. investigated the re-ranking problem from a graph-based perspective and proposed an efficient data structure called ImageWeb to discover the high-level relationships of images [54]. A tradeoff strategy is provided to guide the parameter selection in the online searching process. By sacrificing the initial search accuracy, better search performance is achieved with much lower time complexity compared to the baseline methods and their algorithm is highly scalable. Nemirovskiy and Stoyanov suggested a search pattern based on the rank distributions of the cluster cardinality [55]. Multi-step segmentation performed by means of the recurrent neural network created an image pattern, based on the rank distribution of the brightness cluster cardinalities. The recognition based on the rank distribution allows determining the ND images in grayscale up to the radius of the Gaussian distortion on them. Lei et al. used Radon transform in content-based image representation as it has excellent geometric properties [56]. A family of geometric invariant features based on Radon transform is proposed for near duplicate image detection. NDID methods achieve their goals by measuring the similarity between the features of the query and the target.

An image is represented by a signature and its length gets varied based on the number of patches in the image. Li et al. proposed a visual descriptor named probabilistic center-symmetric local binary pattern (PCSLBP) to depict the patch appearance [57]. Beyond each individual patch, the relationships among the patches are described. A weight is assigned to each patch to identify the image. Given the characteristics of all the patches, the image is represented by a signature. For similarity computation, earth mover's distance is used as it handles variable-length signatures. The patch extraction instability is addressed by allowing many-to-many patch correspondence. Lingxi et al. focussed on fine-grained image search. Fine-grained image search resulted in images that contain the same fine-grained concept with the query [58]. The locality sensitive hashing (LSH) based methods will not generate the buckets of similar sizes, which will reduce the detection effectiveness and efficiency. Yabo et al. proposed load balanced LSH (LBLSH) to produce load-balanced buckets for the hashing process [59]. It reduces the query time and the storage space. Rituparna and Scott evaluated similarity between two images aided by a salient object detection framework [60]. The objective was achieved by incorporating a salient object detection scheme in a sparse representation-based dictionary learning framework. The key aspect in obtaining a more robust similarity measure between images is to extract more meaningful features from the image.

Kim et al. studied near-duplicate image discovery on one billion images which is easily implemented on MapReduce framework [61]. They introduced the seed growing step designed to effectively reduce the number of false positives among cluster seeds. NDI discovery is to detect all clusters composed of images which duplicate at significant regions. Haoran et al. proposed an invariant multi-scale shape descriptor for shape matching and object recognition [62]. This shape feature is invariant to translation, rotation, scaling and can tolerate partial occlusion, articulated variation and intra-class variations. The advantages, disadvantages and suitability of area-based features are tabulated in Table 3.

## 4 Methods Used in Near-Duplicate Retrieval of Images

ND image retrieval is to find all the images that are near duplicates to a particular query. A major challenge in the image retrieval is building effective features that are invariant to a wide range of variations. Near-duplicate retrieval is used for the management of multimedia contents. Search speed is a key factor to judge any near duplicate retrieval algorithm. Not only the retrieval time but also maintaining the relevance of returned results is important in image retrieval. The methods used in the retrieval of near-duplicate images are presented in this section.

Hu et al. proposed a coherent phrase model (CPM) which used the visual phrase of multiple descriptors to characterize every local region to enforce local coherency [63]. It is different from standard Bag of Words (BoW) representation. The two types of phrases proposed are Feature coherent phrase (FCP) and Spatial coherent phrase (SCP). In FCP, every local region is characterized by multiple descriptors of different types. The match of two local regions requires the coherency across different types of features. In SCP, multiple descriptors of a single type of feature are generated between every local region. The match of two regions requires the coherency across different spatial neighbourhoods. Although this model is simple, it improves the matching accuracy by reducing the false matches and additional cost existed for the extraction of multiple features.

The traditional visual vocabulary is created in an unsupervised way by clustering a large number of local features. This is not ideal because it largely ignored the semantic and spatial contexts between local features. Shiliang et al. combined geometric and semantic contexts to generate contextual visual vocabulary [64]. The contextual visual vocabulary is more expensive as the distance computation is time-consuming. Two or more features are combined with other features in the detection of near duplicates. This is known as bundling of features. Bundled features are much more discriminative than a single feature. Wu et al. improved the bundled features with an affine invariant geometric constraint for the retrieval of partial duplicate images [65]. It employs area ratio invariance property of affine transformation to create the affine invariant matrix for bundled visual words. Such affine invariant geometric constraints cope well with flip, rotation or other transformation.

**Table 3** Summary of Features in Area-based Feature Extraction Method

| Features used | Advantages | Disadvantages | Suitability for NDD |
|---|---|---|---|
| Weighted visual words [35] | Each word is assigned a weight de noting the level of similarity | some relevant information is not presented in the set of visual words representation | Clustering of large database of images |
| Contrast context histogram [36] | Computationally fast, requires fewer histogram bins to represent a local region, and has good matching performance. | It only evaluates the intensity differences between the centre pixel and the other pixels in a region which may affect the effectiveness of image matching | More suitable for image matching in real-time applications such as augmented reality |
| HE and WGC [1] | Independent of rotation and scaling trans-formations | Memory and CPU costs are more for large number of images | Image matching |
| PCSLBP [57] | Spatial relationship among the image patches are utilized | Time taken for signature generation is high | Image matching |

The two main strategies for IND retrieval are global feature-based retrieval and local feature-based retrieval. Global features such as color moment and color histogram, although efficient, are sensitive to occlusions and illumination changes. The local region features are robust to lighting, viewpoint and scale changes. However, such methods still suffer from heavy computational cost. Cheng et al. [66] explored local features in spatial-scale space (S-cube) and then formed a global representation for each image. Local features made the model robust to illumination variations, viewpoint and scale changes, as well as geometric transformations, while the final global representation allowed an efficient online retrieval. S-cube has a fast retrieval speed while a high accuracy is preserved. S-cube representation incorporated not only the appearance feature information but also the spatial and scale co-occurrence information. S-cube method is invariant to spatial information and a wider range of scale changes.

Tong et al. proposed a statistical framework for largescale near-duplicate image retrieval using kernel density function [67]. Each image is represented by a kernel density function and the similarity between the query image and a database image is then estimated as the query likelihood. Zhou et al. [68] proposed a matching verification scheme based on BSIFT (binary SIFT) signature. Using the BSIFT, the precision of large-scale image search can be improved by identifying and removing the false-positive matches. Paradowski et al. addressed the problem of largescale near-duplicate image retrieval (NDIR) and proposed a new spatial verification routine [69]. It incorporates neighbourhood consistency; term weighting and it is integrated into the Bhattacharyya coefficient. They give better retrieval quality. The state of the art of technology for NDIR is mostly based on the Bag-of-Visual Words model. However, visual words are straightforward to end in mismatches because of quantization errors of the local features. In order to improve the precision of visual words matching, contextual descriptors are designed to strengthen their discriminative power and measure the contextual similarity of visual words [70]. The contextual descriptor measured the contextual similarity of visual words to discard the mismatches and to reduce the count of candidate images. This descriptor increased the discriminative power of visual words and is robust to image editing operators, such as rotation, scaling and cropping. But it is not robust to perspective transformation of image. It does not work as well on general object retrieval.

Near-duplicate documents are the documents that are created from the original document with some changes. ND document images refer to the images captured from the same document but under different imaging conditions. Document images are the images/scanned copies of documents that may be handwritten or typed documents. In document images, NDD may be used to increase the efficiency of tagging the documents by reducing the need for manual inspection of documents. The entire document (including the text) is considered as image. Document image detection is used in postal automations and digital libraries. Some of the methods used for the detection of near-duplicate document images are presented in this section.

Shiv Vitaladevuni et al. presented an approach to detect near-duplicate document images using SIFT interest point matching [71]. From the set of document images, a database is constructed using the SIFT features extracted from each image. The near duplicates of a query are computed by directly matching the SIFT descriptors with the feature database. Interest point based matching is used for hand and machine written document images for the NDD in Arabic documents in which OCR and text segmentation is difficult. In this approach, 80% of the near duplicate documents are retrieved with low false accepts. Liu et al. proposed document image matching characterized by a graphical perspective [72]. Document images are represented by graphs whose nodes correspond to the objects in the images.

## 5 Near-Duplicate Detection in Image Forensics

Copy-move is one among the most common techniques used for image forgery. In this kind of forgery, one or additional objects in an image are hidden by copying a part and moving it to another place of the same image. Some advanced image editing tools make this type of forgery undetectable by applying a ‘soft’ touch at the edges of the moved part. As the color and texture of the moved part is compatible with those of the copied part, it is very difficult to distinguish between those two parts. Also, two or more identical objects in the same original image make the forgery detection difficult. In copy-move forgery detection, the best performing methods are based on the matching of highly discriminative local features [73]. The discriminability of local features is very less and also, the BOW quantization errors lead to many false local matches that made very difficult to differentiate similar images from copies. Geometric consistency verification reduced the false matches but it neglected the global context information of local features and thus this problem is not solved well. To address this problem, Zhou et al. proposed a global context verification scheme to filter false matches for copy detection [74]. First initial SIFT matches between images are obtained and then the overlapping region-based global context descriptor (OR-GCD) is used for verification of these matches to filter false matches. The OR-GCD has good robustness and efficiency.

Rimba et al. proposed a different kind of scheme by exploiting a group of similar images to verify the source of tampering [75]. By finding the correlation between the reference images and the suspicious image allowed discovering the source of the tampered region. This method relies on the presence of edges. So, the images with no edges are not detected. If the size between the target image and its references are different, then the correlation will be low. Also, if the tampered regions have undergone transformations, then the correlation values will be below the threshold. Irene Amerini et al. presented a SIFT-based forensic method for copy–move attack detection and transformation recovery but it does not investigate the cloned image patch [76]. Ghulam Muhammad et al. proposed a blind copy move image forgery detection method using undecimated dyadic wavelet transform (DyWT) and it is shift invariant [77]. The existing copy-move forgery detection methods either rely on similarity measurements or noise deviation measurements between the parts of an image. Vincent et al. created copy-move forgery by copying and pasting content within the same image [78]. Here, the detection performance is analyzed on the per-image basis and on the per-pixel basis.

I-Chang et al. proposed a forgery detection algorithm to recognize tampered inpainting images [79]. An inpainting image is constructed by filling the target area using the surrounding regions in the same image. Therefore, the key problem is recognizing the fake region and search for similar regions in the faked image. This approach recognized the forged image effectively and identified the forged regions even for the images that have uniform background. In addition to the inpainting type, copy-move forgeries are also detected by this method. But the limitation is in locating forged regions of small sizes. Also, if the forged region is from different sources this method may fail.

## 6 Clustering Methods Used for the NDD

The main aim of clustering is to improve search quality and save storage space. Clustering the web image search engines results is very essential to help users narrow their search. The clustered results are the refined results of the image search. These resultant images will be relevant to the search query. Clustering the near duplicates arising from the results of web image search mainly depends on image features. Image features help to uniquely identify the image from the large dataset. While image search engines return a long list of images, the user may find it difficult to choose his or her interested areas. Generally, the image search results contain multiple topics. Organizing the results into different clusters facilitates users' browsing. Foo and Sinha [11] clustered near duplicate images by combining techniques from invariant image local descriptors and an adaptation of near duplicate text-document clustering techniques. The geometric clue among visual words in an image is computationally expensive and are not considered for clustering.

Some of the online resources such as news and weblogs are used to extract articles and comments are posted related to a popular event. If articles have common parts, then the content of such article is event-relevant leading to near duplicates. Chang et al. proposed an NDD method for finding event-relevant content on the web [81]. The near duplicates related to the same event are checked based on the compact feature of the document. Based on the features, the documents are clustered into event relevant groups. Story clustering is a critical step for news retrieval, topic tracking and summarization. With the overwhelming volume of news videos available today, it becomes necessary to track the development of news stories from different channels, mine their dependencies and organize them in a semantic way. The mining of topic-related stories through clustering on the basis of visual constraints is built on top of text by Wu et al. [82]. Two stories sharing at least one pair of NDK are always placed into the same cluster. Constant-driven coclustering algorithm is employed for mining news topics of varying densities, shapes and sizes. Constraints constructed the association of stories and keyframes; and initialized cluster centers. Also, the retrieval radius is dynamically adjusted to density-reachable stories, which reduce the burden of parameter selection and provide important clues for grouping stories. Constraints may also be outliers and noisy due to the errors in automatic NDK detection.

The ASIFT proposed by Morel et al. viewed the images in different angles by varying the two camera axis orientation parameters, namely, the latitude and the longitude angles, which cannot be done using the SIFT method [83]. This can be done with no computational load. ASIFT outperforms SIFT, maximally stable extremal region (MSER) method, Harris-affine, and Hessian-affine. To reduce the computational cost in partial-duplicate web image search, image features are bundled into local groups. Wu et al. proposed a bundled approach to find the near duplicate images [84]. The SIFT and MSER features were bundled. MSER is a region-based approach, while SIFT is a point feature detection approach. Each group of bundled features becomes more discriminative than a single feature. Kalaiarasi and Thyagharajan clustered the images by bundling the ASIFT and wavelet features of the images [85]. The input to the framework is the results retrieved from the web image search engine. The bundled feature matrix has the relationship among the images with respect to the ASIFT and wavelet feature. Clustering is done using the bundled feature matrix and k-means clustering. Current image search systems use paged image list to display search results. But, the query ambiguity is hard to find search targets in such image lists. The clustering algorithm developed by Ponitz Thomas and Julian Stottinger has a linear runtime and can be carried out in parallel [86]. Features of images with natural repetitive texture become similar to other images and displayed in almost all the search results. This problem is addressed by an asymmetric Hamming distance measurement for bags of visual words. It allows better discrimination power and robust to image transformations such as rotation, cropping, or change of resolution and size.

Zha et al. proposed a group recommendation framework for social groups to share photos [87]. In that framework, pre-built group classifiers predict the group of each photo present in the user's photo collection. Also, representative photos are selected from the collection and accordingly grouping of photos is done. Kalaiarasi and Thyagharajan detected and clustered the ND images based on the visual content present. The supervised and unsupervised clustering is used to form cluster of image [88]. Each cluster would have one image as a representative of that cluster and the other images are called its near duplicates. Clustering of image search results and selection of representative images on large scale image data is done [89]. The system developed by them can organize image search results and help users browse images and find search target. This system applies MapReduce-based image graph construction method and affinity propagation algorithm to generate image clusters and representative images for large-scale image data. This approach does not combine visual and textual image graph. This image grouping system results in the images belonging to multiple clusters. It is an unresolved problem in this work. Partial duplicate image discovery/clustering is defined as finding all the images containing the same objects from a large dataset. The challenge of partial duplicate web image discovery/clustering is that the target images may contain many variations, such as color, scale, illumination; as well as viewpoint changes, partial occlusion, near-duplicate and affine transformations. Zhang and Qiu improved the performance of large scale partial duplicate image discovery and clustering by encoding the spatial geometric information of bag of visual words (BOW) [90].

## 7 Data Bases Used for NDD Evaluation

The widely used databases to evaluate the near-duplicate detection of images are discussed in this section. **Corel Photo CD Collection: [91]** Each image under this set is altered using the following list of transformations – format change (1), colorsize (3), contrast (2), severe contrast (2), crop (4), severe crop (3), despeckle (1), frame (4), rotate (3), scale-up (3), scale-down (3), saturation (5), intensity (4), severe intensity (2), rotate + crop (3), rotate + scale (3), shear (4). This results in 50 alterations [11, 80]. **MIRFlickr1 M and MIRFlickr60K: [92]** MIRFlickr1M has 1 million distractor images mostly related to holidays of humans. It also includes low resolution images. MIRFlickr60K has 67,714 images.

MIRFLICKR-1M and 2008 - MIRFLICKR25000 [93]: MIRFlickr25000 has 25000 images and MIRFlickr1M has one million distractor images which are used to evaluate large scale image search. Compared to INRIA, the Flickr datasets are slightly biased, because they include low resolution images and more photos of humans. **Oxford5K Dataset: [94]** The dataset has 5062 images of Oxford buildings collected from Flickr by searching Oxford landmarks. [45] introduce a novel quantization method based on randomized trees and evaluate it using Oxford dataset. They show that the quality of image retrieval depends on the quantization.

**MM270K Dataset: [95]** This set contains about 18,000 images with very diverse content, such as animals, landscapes, people, etc. These images have undergone 40 transforms ([9, 63]).

**Columbia NDI Database: [96]** The database contains much greater variation in spatial translations and scale variations. There are 150 ND pairs (300 images) and 300 ND images [18, 38, 63, 98].

**CityU Dataset: [97]** The set consists of 29 news topic covering 805 stories and resulting in 7006 shots [38, 63]. **NTU Dataset: [99]** There are 150 ND pairs (300 images) and 300 ND images [63].

**UKBench Dataset: [100]** The dataset contains 10,200 images where every four images capture a single object with the different viewpoint, illumination conditions and scale. There is a total of 2550 objects such as CD covers, etc. [102].

**INRIA Dataset: [101]** The dataset contains 1491 images in 500 groups with a variety of scene types such as natural, man-made, water, sky, etc. These images were modified by applying different types of transformations like image rotation, image resizing, JPEG compression ranging from JPEG3 to JPEG 75, image cropping ranging from 5% to 80% of the image surface and by applying strong transformations such as print and scan, paint, change in contrast, perspective effect, blur, very strong crop and so on [53].

**California ND Dataset: [103]** This dataset has 701 photos of a user's personal photo collection. Many challenging nonidentical near-duplicates of real-world scenes are provided in this collection. Detecting NDs is a first step required in managing photo collections. This dataset helps the researchers to test their NDD algorithms with real photos [2]. The first 604 photos were consecutively selected from the beginning of the collection. The remaining 97 photos consist of pairs of interesting cases from the same collection that is not present among the first 604 photos. The images undergone transformations such as burst shots (344 photos), moving background shots (149 photos), show/performance shots (54 photos), group photos (17 photos), panorama shots (8 photos), exposure/brightness difference (58 photos), viewpoint difference/zooming (36 photos), focus change (22 photos), white balance difference (8 photos). The case of burst, performance and panorama shots have not been considered in any other datasets.

**COVERAGE Database: [104]** The COVERAGE (COpymoVeforgERydAtabase with similar but GenuineobjEcts) contains 100 original images and 100 forged images [3]. In this database, the forged/near-duplicate images have been obtained by performing transformations such as Translation, Scaling, Rotation, Free-Form, Illumination and Combination. Out of 100 near-duplicate images, 16 images fall under translation transformation, 16 images under scaling, 16 images under rotation, 16 images under free-form, 16 images under illumination and 20 images under combination transformation [3].

**CoMoFoD Database: [105]** The CoMoFoD (Copy-Move Forgery Detection) has 200 small images. In this database, the forged/near-duplicate images have been obtained by performing transformations such as Translation, Scaling, Rotation, Free-Form, Illumination and Combination [106]. Also subjected to the following post-processing methods to derive many more duplicate images and to make the forgery perfect - Brightness Change (3), Contrast adjustment (3), Color reduction (3), Image blurring (3), JPEG compression (9), Noise addition (3). These 24 transformations are done for both the original and forged image resulting in 48 transformations [107].

**CASIA Database: [107]** The CASIA V1.0 dataset has 800 authentic images mostly collected from Corel image set and it also has 921 spliced color images in JPEG format. Image splicing is the process of cutting certain regions of one image and pasting them on the same or other images without performing post-processing. The CASIA V2.0 database consists of 7491 authentic more realistic images and 5123 tampered color images with post-processed boundaries of spliced regions.

**MICC-F220 and MICC-F2000: [108]** Irene Amerini [76] created MICC-F220 a database with 220 images and MICCF2000 a database with 2000 images. MICC-F220 consists of 110 original images and 110 tampered images and the image resolution varies from 722 × 480 to 800 × 600. The average size of the forged patch is 1.2% of the image size. MICC-F2000 has 1300 original images and 700 tampered images with the resolution of 2048 × 1536 pixels. MICCF600 dataset created by the same authors has 440 original images, 160 tampered images and 160 ground truth images.

The CoMoFoD, Manipulation (Manip) and GRIP datasets [3] provide both the forged and original images. Among these, only CoMoFoD considers complex manipulations for forged image analysis. From these discussions, the CoMoFoD database is found to be more suitable for justifying the near-duplicate detection method as there are various complex transformations. If the proposed method detects near duplicate images properly in this database, then that method will be more appropriate and best suited for all types of duplication/forgeries detection. The performance of various feature extraction methods tested with different near-duplicate datasets is compared in Table 4.

## 8 Applications of Near-Duplicate Detection of Images

Some of the applications developed by the researchers based on NDD are given below

1. Identifying and eliminating ND spam reviews, thereby providing a summary of the trusted reviews for customers to make buying decisions [111]
2. Fabricated Picture Detection [112]
3. Zhang et al. presented a face annotation system to collect and label celebrity faces automatically from the web [113]
4. Face matching system for post-disaster family reunification [114]
5. Logo recognition technique based on feature bundling [115]
6. Logo search is required in many real-world applications [116]
7. Building a digital library and postal automation [57]
8. Hand gesture recognition [63]
9. Resource utilization and traffic alleviation in many network architectures and leveraging in-network storage for various content-centric sessions [117]
10. Visual media intelligence in detecting similar visual content published on the web in a short period of time [118]

## 9 Challenges and Future Research Directions

A major challenge for the NDD is the runtime performance. If the runtime is more, then it leads to a huge waste of computational resources. The detection efficiency of traditional approaches is not satisfactory because the peculiarity of the NDD is very huge and some “hot spot” images have too many duplicates while some have very few. On the whole, the present systems fail to produce good results when the duplicates are created by large rotation, large scaling, many simultaneous image manipulations, change in the viewpoint and transformations in image tampering. These systems also suffer from the huge waste of network resources, high computational complexity, high false identification and low accuracy. The main challenge in this area is to find out the original image from the set of similar images. Also, due to the presence of NDs, the search time increases in web search engines. Even though many researchers worked on the scalability and reliability issue, still those issues should be addressed and improved a lot.

## 10 Conclusion

The search engines face a problem leading to the fact that the search results are of less relevance to the user due to the presence of near-duplicate images. Since the digital content is widespread and also easily redistributable, the presence of near duplicates affects the performance of the search engines critically. In this paper, the state-of-the-art approaches of detecting near-duplicate images and their applications are reviewed along with the methodologies used. The main challenges in this field are discussed. This review provides research directions to the fellow researchers who are interested to work in this field. Also, the survey concentrated only on the feature extraction task in computer vision system. There are many more tasks and techniques to do research.

**Table 4** Performance comparison of features extraction methods on different datasets

| ND dataset | Author who used | Feature extraction | Precision | Recall | Accuracy |
|---|---|---|---|---|---|
| Corel photo CD collection | Foo and Sinha [12] | RBV | 99 | 97 | – |
| | Foo et al. [80] | Text-oriented clustering algorithm | 76.9 | 77.1 | – |
| Flickr | Wu et al. [34] | Query oriented subspace shifting | 96.85 | 90.34 | – |
| | Li et al. [53] | Visual word model | 83 | 73.6 | – |
| | Chu and Lin [109] | NDD without filtering | 33 | 41 | – |
| | | NDD with filtering | 57 | 20 | – |
| MM270K | Ke et al. [9] | Parts-based approach | 88.78 | 96.78 | – |
| | Hu et al. [63] | Locality Sensitive Hashing | 90 | 90 | – |
| Columbia | Xu et al. [98] | Spatially Aligned Pyramid Matching | 82.3 | 82.3 | – |
| | Wang et al. [18] | Keypoint based Near-duplicate detection | 98.33 | 82.33 | – |
| | Hu et al. [63] | Feature Coherent Phrase | – | – | 83.3 |
| | | Spatial Coherent Phrase | – | – | 81.3 |
| | Zhao and Ngo [37] | Scale Rotation-Pattern Entropy | 100 | 82.4 | – |
| | | Pattern Entropy | 97.7 | 81 | – |
| | | RANSAC-based Cardinality Threshold | 91.2 | 79.1 | – |
| | | Cardinality Threshold | 96.1 | 70.9 | – |
| | | Visual Keyword | 59.3 | 59.3 | – |
| | | Color Moment | 40 | 40 | – |
| CityU | Zhao and Ngo [37] | Scale Rotation-Pattern Entropy | 91 | 78.4 | – |
| | | Pattern Entropy | 90.7 | 76.9 | – |
| | | RANSAC-based Cardinality Threshold | 87.3 | 74.8 | – |
| | | Cardinality Threshold | 86.8 | 74 | – |
| | | Visual Keyword | 37.2 | 37.2 | – |
| | | Color Moment | 19.2 | 19.2 | – |
| | Hu et al. [63] | Feature Coherent Phrase | 74.9 | 74.9 | – |
| NTU | Hu et al. [63] | Feature Coherent Phrase | 96 | 96 | – |
| | | Spatial Coherent Phrase | 96 | 96 | – |
| UKBench | Battiato et al. [102] | Codebook alignment for NDD | 73.4 | 73.4 | – |
| Copydays | Li et al. [53] | Visual word model | 84.6 | 76.3 | – |
| California ND | Eshkol et al. [110] | MPEG-7 Color structure descriptor | 54.4 | 78.2 | – |

## Compliance with Ethical Standards

**Conflict of interest** The authors have no conflict of interest in publishing this Author's version (first version) of the article in ArXiv

## References

1. Jegou H, Douze M, Schmid C (2008) Hamming embedding and weak geometry consistency for large scale image search. In: Proceedings of the 10th European conference on computer vision. https ://doi.org/10.1007/978-3-540-88682 -2_24
2. Jinda-Apiraksa A, Vonikakis V,Winkler S (2013) California-ND: an annotated dataset for near- duplicate detection in personal photo collections. In: Proceedings in 5th international workshop on quality of multimedia experience (QoMEX), Klagenfurt, Austria. https ://doi.org/10.1109/QoMEX .2013.66032 27
3. Wen B, Zhu Y, Subramanian R, Ng TT, Shen X, Winkler S (2016) COVERAGE—a novel database for copy-move forgery Detection. In: IEEE international conference on image processing (ICIP), Phoenix, AZ, USA, pp 161–165. https ://doi. org/10.1109/ICIP.2016.75323 39
4. Giuseppe Toys dataset. http://www.vision .caltech.edu/pmoree ls / Datas ets/Giuse ppe_Toys_03/. Accessed 14 Jul 2018
5. Chum O, Philbin J, Isard M, Zisserman A (2007) Scalable near identical image and shot detection. In: Proceedings of the 6th ACM international conference on image and video retrieval, pp 549–556. https ://doi.org/10.1145/12822 80.12823 59
6. Amruta Landge, Mane Pranoti (2016) Near duplicate image matching techniques. IEEE Int Conf Inf Commun Embed Syst ICICES. https ://doi.org/10.1109/ICICE S.2016.75188 63
7. Thajeel SA, Sulong GB (2013) State of the art of copy-move forgery detection techniques: a review. Int J Comput Sci Issues 10(6):174–183. https ://doi.org/10.1109/ICCS1 .2017.83259 63
8. Spyrou E, Mylonas P (2016) A survey on Flickr multimedia research challenges. Eng Appl Artif Intell. https ://doi. org/10.1016/j.engap pai.2016.01.006

9. Ke Y, Sukthankar R, Huston L (2004) Efficient near-duplicate detection and sub-image retrieval. ACM Multimed 4(1):5
10. Chen L, Fred S (2006) Comparison of near-duplicate image matching. In: Proceedings of the 3rd European conference on visual media production. https ://doi.org/10.1049/cp:20061 969
11. Foo JJ, Sinha R (2007a) Pruning SIFT for scalable near-duplicate image matching. In: Proceedings of the eighteenth conference on Australasian database, vol 63, pp 63–71. Australian Computer Society, Inc
12. Foo JJ, Sinha R (2007b) Using redundant bit vectors for nearduplicate image detection. In: DASFAA, pp 472–484. https: //doi. org/10.1007/978-3-540-71703 -4_41
13. Bay H, Ess A, Tuytelaars T, Van Gool L (2008) Speeded-up robust features (SURF). Comput Vis Image Underst 110(3):346–359
14. Rublee E, Rabaud V, Konolige K, Bradski GR (2011) ORB: an efficient alternative to SIFT or SURF. In: ICCV, vol 11, no 1, p 2
15. Cao, Y, Zhang H, Gao Y, Guo J (2010) An efficient duplicate image detection method based on Affine-SIFT feature. In: 3rd IEEE international conference on broadband network and multimedia technology (IC-BNMT), pp 794–797. https ://doi. org/10.1109/ICBNM T.2010.57051 99
16. Yu Guoshen, Morel Jean-Michel (2011) ASIFT: an algorithm for fully affine invariant comparison. Image Process Line 1:11–38. https ://doi.org/10.5201/ipol.2011
17. Lindeberg T (2013) Scale selection properties of general-ized scale-space interest point detectors. J Math Imaging Vis 46(2):177–210
18. Wang Y, Hou Z, Leman K (2011) Keypoint-based near-duplicate images detection using affine invariant feature and color matching. In: International conference in acoustics, speech and signal processing (ICASSP), pp 1209–1212. https ://doi.org/10.1109/ ICASS P.2011.59466 27
19. Tareen SAK, Saleem Z (2018) A comparative analysis of sift, surf, kaze, akaze, orb, and brisk. In: 2018 International conference on computing, mathematics and engineering technologies (iCoMET), pp 1–10. IEEE
20. Foo JJ, Zobel J, Sinha R, Tahaghoghi SM (2007b) Detection of near-duplicate images for web search. In: Proceedings of the 6th ACM international conference on image and video retrieval, pp 557–564. https ://doi.org/10.1145/12822 80.12823 60
21. Zhang S, Tian Q, Lu K, Huang Q, Gao W (2013) Edge-SIFT: discriminative binary descriptor for scalable partial-duplicate mobile search. IEEE Trans Image Process 22(7):2889–2902. https ://doi.org/10.1109/TIP.2013.22516 50
22. Muresan RC (2003) Pattern recognition using pulse-coupled neural networks and discrete Fourier transforms. Neurocomputing 51:487–493. https ://doi.org/10.1016/S0925 -2312(02)00727 -0
23. Zhang YD, Wu LN (2008) Pattern recognition via PCNN and Tsallis entropy. Sensors 8(11):7518–7529. https ://doi. org/10.3390/s8117 518
24. Ma Y, Wang Z, Wu C (2006) Feature extraction from noisy image using PCNN. In: Proceedings of the international conference on information acquisition, pp 808–813. https ://doi.org/10.1109/ icia.2006.30583 4
25. Xiadong Gu (2008) Feature extraction using unit-linking pulse coupled neural network and its applications. Neural Process Lett
27:25–41. https ://doi.org/10.1007/s1106 3-007-9057-6
26. Forgac R, Mokris I (2008) Feature generation improving by optimised PCNN. In: Proceedings of 6th international symposium on applied machine intelligence and informatics, pp 203–207. https ://doi.org/10.1109/sami.2008.44691 66
27. Hoang Trong-Thuc, Nguyen Ngoc-Hung, Nguyen Xuan-Thuan, Bui Trong-Tu (2012) A real-time image feature extraction using pulse-coupled neural network. Int J Emerg Trends Technol Comput Sci IJETICS 1(3):117–185
28. Mohammed MM, Abdelhalim MB, Badr A (2014) An optimised PCNN for image classification. In: 10th international computer engineering conference (ICENCO), Giza, pp 16–20. https ://doi. org/10.1109/ICENC O.2014.70504 25
29. An L, Yin G, Gao X (2013) Graph matching with geometric constraints for near-duplicated image retrieval. In: Proceedings of the fifth international conference on internet multimedia computing and service, pp 174–177. ACM. https ://doi.org/10.1145/24997 88.24998 47
30. Zhang Y, Zhang Y, Sun J, Li H, Zhu Y (2018) Learning near duplicate image pairs using convolutional neural networks. Int J Perform Eng 14(1):168

31. Vonikakis V, Chrysostomou D, Kouskouridas R, Gasteratos A (2012) Improving the robustness in feature detection by local contrast enhancement. In: Imaging systems and techniques (IST), IEEE international conference, pp 158–163. https ://doi. org/10.1109/ist.2012.62954 82
32. Vonikakis V, Chrysostomou D, Kouskouridas R, Gasteratos A (2013) A biologically inspired scale-space for illumination invariant feature detection. Meas Sci Technol 24(7):074024. https ://doi.org/10.1088/0957-0233/24/7/07402 4
33. Zhuang D, Zhang D, Li J, Tian Q (2015) Binary feature from intensity quantization and weakly spatial contextual coding for image search. Inf Sci 302:94–107. https ://doi.org/10.1016/j. ins.2014.08.064
34. Wu L, Liu J, Yu N, Li M (2008) Query oriented subspace shifting for near-duplicate image detection. In: IEEE international conference on multimedia and expo, pp 661–664. https ://doi. org/10.1109/icme.2008.46075 21
35. Chum O, Philbin J, Zisserman A (2008) Near duplicate image detection: min-hash and tf-idf weighting. BMVC 810:812–815. https ://doi.org/10.5244/C.22.50
36. Huang Chun-Rong, Chen Chu-Song, Chung Pau-Choo (2008) Contrast context histogram-An efficient discriminating local descriptor for object recognition and image matching. Pattern Recognit 41:3071–3077. https ://doi.org/10.1016/j.patco g.2008.03.013
37. Zhao WL, Ngo CW (2009) Scale-rotation invariant pattern entropy for keypoint-based near- duplicate detection. IEEE Trans Image Process 18(2):412–423. https ://doi.org/10.1109/ tip.2008.20089 00
38. Battiato S, Farinella GM, Guarnera GC, Meccio T, Puglisi G, Ravì D, Rizzo R (2010) Bags of phrases with codebooks alignment for near duplicate image detection. In: Proceedings of the 2nd ACM workshop on multimedia in forensics, security and intelligence, pp 65–70. https: //doi.org/10.1145/1877972.187799 1
39. Sluzek A, Paradowski M, Duanduan Y (2010) Detection and segmentation of near-duplicate fragments in random images. In: Control automation robotics and vision (ICARCV), 11th international conference, pp 1161–1166. https: //doi.org/10.1109/ICARC V.2010.57072 94
40. Cho A, Yang WK, Oh WG, Jeong DS (2010) Concentric circle based image signature for near-duplicate detection in large databases. ETRI J 32(6):871–880. https ://doi.org/10.4218/etrij .10.0109.0623
41. Ferrari V, Tuytelaars T, Van Gool L (2004) Simultaneous object recognition and segmentation by image exploration. In: European conference on computer vision, pp 40–54. Springer, Berlin, Heidelberg. https ://doi.org/10.1007/978-3-540-24670- 1_4
42. Jegou Herve, Douze Matthijs, Schmid Cordelia (2010) Improving bag-of-features for large scale image search. Int J Comput Vis 87(3):316–336. https ://doi.org/10.1007/s1126 3-009-0285-2
43. Nister D, Stewenius H (2006) Scalable recognition with a vocabulary tree. In: Proceedings of IEEE conference on computer vision and pattern recognition, pp 2161–2168. https ://doi. org/10.1109/CVPR.2006.264
44. Philbin J, Chum O, Isard M, Sivic J, Zisserman A (2007) Object retrieval with large vocabularies and fast spatial matching. In: Proceedings of the IEEE conference on computer vision and pattern recognition, pp 1–8. https ://doi.org/10.1109/ CVPR.2007.38317 2
45. Philbin J, Chum O, Isard M, Sivic J, Zisserman A (2008) Lost in quantization: improving particular object retrieval in large scale image databases. In: Proceedings of IEEE conference on computer vision and pattern recognition, pp 1–8. https ://doi. org/10.1109/cvpr.2008.45876 35
46. Xie H, Gao K, Zhang Y, Tang S, Li J, Liu Y (2011) Efficient feature detection and effective post-verification for large scale nearduplicate image search. IEEE Trans Multimed 13(6):1319–1332. https ://doi.org/10.1109/tmm.2011.21672 24
47. Satoh SI (2011) Simple low-dimensional features approximating NCC-based image matching. Pattern Recognit Lett 32(14):1902–1911. https ://doi.org/10.1016/j.patre c.2011.07.027
48. Das SN, Mathew M, Vijayaraghavan PK (2012) An efficient approach for finding near duplicate web pages using minimum weight overlapping method. In: Ninth IEEE international conference in information technology: new generations (ITNG), pp 121–126. https ://doi.org/10.1109/ITNG.2012.168
49. Dong W, Wang Z, Charikar M, Li K (2012) High-confidence near-duplicate image detection. In: Proceedings of the 2nd ACM international conference on multimedia retrieval, p 1. https: //doi. org/10.1145/23247 96.23247 98

50. Bueno L, Valle E, da Torres SR (2012) Bayesian approach for near-duplicate image detection. In: Proceedings of 2nd ACM international conference on multimedia retrieval, pp 15. https :// doi.org/10.1145/23247 96.23248 15
51. Rakthanmanon T, Zhu Q, Keogh EJ (2012) Efficiently finding near duplicate figures in archives of historical documents. J Multimed 7(2):109–123
52. Li P, Hanbing YAN, Gang CUI, Yuejin DU (2012) Near-duplicate image identification with geometric consistency verification. J Comput Inf Syst 9:3593–3603
53. Li Z, Feng X (2013) Near duplicate image detecting algorithm based on bag of visual word model. J Multimed 8(5):557–565. https ://doi.org/10.4304/jmm.8.5.557-564
54. Xie L, Tian Q, Zhou W, Zhang B (2014) Fast and accurate near-duplicate image search with affinity propagation on the ImageWeb. Comput Vis Image Underst. https: //doi.org/10.1016/j. cviu.2013.12.011
55. Nemirovskiy VB, Stoyanov AK (2014) Near-duplicate image recognition based on the rank distribution of the brightness clusters cardinality. Comput Opt 38(4):811–817. https ://doi. org/10.18287 /0134-2452-2014-38-4-811-817
56. Lei Y, Zheng L, Huang J (2014) Geometric invariant features in the radon transform domain for near-duplicate image detection. Pattern Recognit 47(11):3630–3640. https: //doi.org/10.1016/j. patco g.2014.05.009
57. Li L, Yue L, Ching YS (2015) Variable-length signature for near-duplicate image matching. IEEE Trans Image Process 24(4):1282–1296. https ://doi.org/10.1109/TIP.2015.24002 29
58. Xie L, Wang J, Zhang B, Tian Q (2015) Fine-grained image search. IEEE Trans Multimed 17(5):636–647. https ://doi. org/10.1109/tmm.2015.24085 66
59. Fan Y, Xing J, Hu W (2015) Load-balanced locality-sensitive hashing: a new method for efficient near duplicate image detection. In: IEEE international conference in image processing (ICIP), pp 53–57. https ://doi.org/10.1109/ICIP.2015.73507 58
60. Sarkar R, Acton ST (2016) SLIDE: saliency guided image dictionary and image similarity evaluation. In: ICIP 2016, pp 216–220. https ://doi.org/10.1109/icip.2016.75323 50
61. Kim S, Wang XJ, Zhang L, Choi S (2015) Near duplicate image discovery on one billion images. In: IEEE winter conference in applications of computer vision (WACV), pp 943–950. https: // doi.org/10.1109/wacv.2015.130
62. Haoran Xu, Yang Jianyu, Yuan Junsong (2016) Invariant multiscale shape descriptor for object matching and recognition. IEEE Int Conf Image Process ICIP. https ://doi.org/10.1109/ ICIP.2016.75324 36
63. Hu Y, Cheng X, Chia LT, Xie X, Rajan D, Tan AH (2009) Coherent phrase model for efficient image near-duplicate retrieval. IEEE Trans Multimed 11(8):1434–1445. https: //doi. org/10.1109/TMM.2009.20326 76
64. Zhang Shiliang, Tian Qi, Hua Gang, Zhou Wengang, Huang Qingming, Li Houqiang, Gao Wen (2011) Modeling spatial and semantic cues for large-scale near-duplicated image retrieval. Comput Vis Image Underst 115:403–414. https: //doi. org/10.1016/j.cviu.2010.11.003
65. Wu Z, Xu Q, Jiang S, Huang Q, Cui P, Li L (2010) Adding affine invariant geometric constraint for partial-duplicate image retrieval. In: 20th IEEE international conference in pattern recognition (ICPR), pp 842–845. https: //doi.org/10.1109/ ICPR.2010.212
66. Cheng X, Hu Y, Chia LT (2011) Exploiting local dependencies with spatial-scale space (s-cube) for near-duplicate retrieval. Comput Vis Image Underst 115(6):750–758. https ://doi. org/10.1016/j.cviu.2011.02.003
67. Tong Wei, Li Fengjie, Jin Rong, Jain Anil (2012) Large-scale near-duplicate image retrieval by kernel density estimation. Int J Multimed Inf Retr 1:45–58. https ://doi.org/10.1007/s1373 5-012-0012-6
68. Zhou W, Li H, Lu Y, Wang M, Tian Q (2015) Visual word expansion and BSIFT verification for large-scale image search. Multimed Syst 21(3):245–254. https ://doi.org/10.1007/s0053 0-013-0330-4
69. Paradowski M, Durak M, Broda B (2014) Bag of words-quality issues of near-duplicate image retrieval. Mach Gr Vis 23(1):83–96
70. Yao J, Yang B, Zhu Q (2015) Near-duplicate image retrieval based on contextual descriptor. IEEE Signal Process Lett 22(9):1404–1408. https ://doi.org/10.1109/LSP.2014.23777 95

71. Vitaladevuni S, Choi F, Prasad R, Natarajan P (2012) Detecting near-duplicate document images using interest point matching. In: 21st IEEE international conference on pattern recognition (ICPR), pp 347–350
72. Liu L, Lu Y, Suen CY (2014) Near-duplicate document image matching: a graphical perspective. Pattern Recognit 47(4):1653–1663. https ://doi.org/10.1016/j.patco g.2013.11.006
73. Picard D (2016) Preserving local spatial information in image similarity using tensor aggregation of local features. In: ICIP, pp 201–205. https ://doi.org/10.1109/ICIP.2016.75323 47
74. Zhou Z, Wang Y, Wu QJ, Yang CN, Sun X (2017) Effective and efficient global context verification for image copy detection. IEEE Trans Inf Forensics Secur. https ://doi.org/10.1109/ tifs.2016.26010 65
75. Ciptasari Rimba Whidiana, Rhee Kyung Hyune, Sakurai Kouichi (2013) Exploiting reference images for image splicing verification. Digit Investig 10(2013):246–258. https: //doi.org/10.1016/j. diin.2013.06.014
76. Amerini I, Ballan L, Caldelli R, Del Bimbo A, Serra G (2011) A SIFT-based forensic method for copy–move attack detection and transformation recovery. IEEE Trans Inf Forensics Secur 6(3):1099–1110. https ://doi.org/10.1109/tifs.2011.21295 12
77. Muhammad G, Hussain M, Bebis G (2012) Passive copy move image forgery detection using undecimated dyadic wavelet transform. Digit Investig 9(1):49–57. https ://doi.org/10.1016/j. diin.2012.04.004
78. Christlein Vincent, Riess Christian, Jordan Johannes, CorinnaRiess Elli Angelopoulo (2012) An evaluation of popular copymove forgery detection approaches. IEEE Trans Inf Forensics Secur 7(6):1841–1854. https: //doi.org/10.1109/TIFS.2012.22185 97
79. Chang IC, Yu JC, Chang CC (2013) A forgery detection algorithm for exemplar- based inpainting images using multiregion relation. Image Vis Comput 31(1):57–71. https ://doi. org/10.1016/j.imavi s.2012.09.002
80. Foo JJ, Zobel J, Sinha R (2007a) Clustering near-duplicate images in large collections. In: Proceedings of the international workshop on multimedia information retrieval, pp 21–30. ACM. https ://doi.org/10.1145/12900 82.12900 89
81. Chang HC, Wang JH, Chiu CY (2007) Finding event-relevant content from the web using a near-duplicate detection approach. In: Proceedings of the IEEE/WIC/ACM international conference on web intelligence, pp 291–294. https ://doi.org/10.1109/ WI.2007.58
82. Wu X, Ngo CW, Hauptmann AG (2008) Multimodal news story clustering with pairwise visual near-duplicate constraint. IEEE Trans Multimed 10(2):188–199. https ://doi.org/10.1109/ tmm.2007.91177 8
83. Jean-Michel M, Yu G (2009) ASIFT: a new framework for fully affine invariant image comparison. SIAM J Imaging Sci 2:438–469. https ://doi.org/10.1137/08073 2730
84. Wu Z, Ke Q, Isard M, Sun J (2009) Bundling features for large scale partial-duplicate web image search. In: IEEE conference in computer vision and pattern recognition, CVPR, pp 25–32. https ://doi.org/10.1109/cvpr.2009.52065 66
85. Kalaiarasi G, Thyagharajan KK (2017) Clustering of near duplicate images using bundled features. Clust Comput J. https ://doi. org/10.1007/s1058 6-017-1539-3
86. Ponitz T, Stottinger J (2010) Efficient and robust near-duplicate detection in large and growing image data-sets. In: Proceedings of the 18th ACM international conference on multimedia, pp 1517–1518. https ://doi.org/10.1145/18739 51.18742 68
87. Zha ZJ, Tian Q, Cai J, Wang Z (2013) Interactive social group recommendation for Flickr photos. Neurocomputing 105:30–37. https ://doi.org/10.1016/j.neuco m.2012.06.039
88. Kalaiarasi G, Thyagharajan KK (2013) Visual content based clustering of near duplicate web search images. In: The proceeding of IEEE international conference on green computing, communication and conservation of energy (ICGCE), India, pp 767–71. https ://doi.org/10.1109/icgce .2013.68235 37
89. Hsieh L-C, Wu G-L, Hsu Y-M, Hsu W (2014) Online image search result grouping with MapReduce-based image clustering and graph construction for large-scale photos. J Vis Commun Image Represent 2:384–395. https: //doi.org/10.1016/j.jvcir .2013.12.010
90. Zhang Q, Qiu G (2015) Geometric consistent tree partitioning min-hash for large-scale partial duplicate image discovery. In: IEEE international conference in multimedia big data (BigMM), pp 220–227. https ://doi.org/10.1109/bigmm .2015.38

91. Corel Photo CD Collection: http://apps.corel. com/dell/paintshop / uk/photo _album _6/downl oad.html. Accessed 8 May 2018
92. MIRFlickr dataset: http://press .liacs .nl/mirfl ickr/. Accessed 8 May 2018
93. Huiskes MJ, Lew MS (2008) The MIR Flickr retrieval evaluation. In: ACM international conference on multimedia information retrieval (MIR'08), Vancouver, Canada. https ://doi. org/10.1145/14600 96.14601 04
94. OXFORD Building dataset http://www.robot s.ox.ac.uk/~vgg/ data/oxbui lding s/. Accessed 30 April 2018
95. MM270K Dataset http://www.cs.cmu.edu/~yke/retri eval
96. Columbia NDI Database: http://www.ee.columb ia.edu/ln/dvmm/ downl oads/AuthS plice dData Set/AuthS plice dData Set.htm. Accessed 8 May 2018
97. CityU Dataset: http://vireo .cs.cityu .edu.hk/resea rch/ndk/ndk. html. Accessed 8 May 2018
98. Xu D, Cham TJ, Yan S, Duan L, Chang SF (2010) Near duplicate identification with spatially aligned pyramid matching. IEEE Trans Circuits Syst Video Technol 20(8):1068–1079
99. NTU Dataset: http://claren celiang.com/datase t/NTU_Data set.zi p. Accessed 8 May 2018
100. UKBench Dataset: http://www.vis.uky.edu/~stewe /ukben ch. Accessed on 8 May 2018
101. INRIA dataset http://lear.inrial pes.fr/~jegou/data.ph p. Accessed 8 May 2018
102. Battiato S, Farinella GM, Puglisi G, Ravì D (2014) Aligning codebooks for near duplicate image detection. Multimed Tools Appl 72(2):1483–1506. https ://doi.org/10.1007/s1104 2-013-1470-4
103 . California-ND Dataset: http://vintag e.winklerbros .net/cali fornia ND.html. Accessed 3 May 2018
104. COpy-moVe forgERy dAtabase with similar but Genuine objects (COVERAGE).https: //github.com/wenbih a/cove rag e.Accessed 4 May 2018
105. Copy move forgery detection(CoMoFoD). http://www.vcl.fer.hr/ comof od. Accessed 4 May 2018
106. Dijana T, Zupancic I, Grgic S, Grgic M (2013) CoMoFoD: new database for copy- move forgery detection. In: Proceedings in 55th international symposium (ELMAR), pp 49–54
107 . CASIA Database: http://forens ics.idealtest.or g. Accessed 8 May 2018
108. MICC-F220 and MICC-F2000: http://lci.micc.unifi .it/ labd/2015/01/copy-move-forge ry-detec tion-and-local izati on/. Accessed 8 May 2018
109. Chu WT, Lin CH (2010) Consumer photo management and browsing facilitated by near-duplicate detection with feature filtering. J Vis Commun Image Represent 21(3):256–268. https :// doi.org/10.1016/j.jvcir .2010.01.006
110. Eshkol A, Grega M, Leszczuk M, Weintraub O (2014) Practical application of near duplicate detection for image database. In: International conference on multimedia communications, services and security, pp 73–82, Springer, Cham. https ://doi. org/10.1007/978-3-319-07569 -3_6
111. Algur SP, Patil AP, Hiremath PS, Shivashankar S (2010) Conceptual level similarity measure based review spam detection. In: Signal and image processing (ICSIP), IEEE international conference, pp 416–423. https ://doi.org/10.1109/ICSIP .2010.56975 09
112. Tang X (2012) Book retrieval based on near-duplicate image matching. In: Fuzzy systems and knowledge discovery (FSKD), 9th IEEE international conference, pp 2616–2619. https ://doi. org/10.1109/FSKD.2012.62337 92
113. Zhang X, Zhang L, Wang XJ, Shum HY (2012) Finding celebrities in billions of web images. IEEE Trans Multimed 14(4):995–1007. https ://doi.org/10.1109/TMM.2012.21861 21
114. Borovikov E, Vajda S, Lingappa G, Antani S, Thoma G (2013) Face matching for post-disaster family reunification. In: IEEE international conference on healthcare informatics (ICHI), pp 131–140. https ://doi.org/10.1109/ICHI.2013.23
115. Romberg S, Lienhart R (2013) Bundle min-hashing for logo recognition. In: Proceedings of the 3rd ACM conference on international conference on multimedia retrieval, pp 113–120. https ://doi.org/10.1145/24614 66.24614 86
116. Xie L, Tian Q, Zhang B (2014) Max-SIFT: Flipping invariant descriptors for web logo search. In: IEEE international conference in image processing (ICIP), pp 5716–5720. https :// doi.org/10.1109/ICIP.2014.70261 56

117. Cui H, Yuan X, Zheng Y, Wang C (2016) Enabling secure and effective near-duplicate detection over encrypted in-network storage. IEEE INFOCOM—the 35th annual international conference in computer communications, pp 1–9. https ://doi.org/10.1109/
INFOC OM.2016.75243 46
118. Gadeski E, Le Borgne H, Popescu A (2017) Fast and robust duplicate image detection on the web. Multimed Tools Appl 76(9):11839–11858. https: //doi.org/10.1007/s11042-016-3619- 4